\documentclass[11pt]{article}

\usepackage[final]{acl}

\usepackage{wrapfig}

\usepackage{amsmath}
\usepackage{amssymb}
\usepackage{amsfonts}                               
\usepackage{amsthm}
\usepackage[mathcal]{eucal}
\usepackage{mathrsfs}
\usepackage{bm}     

\usepackage{cleveref}
\usepackage{subfigure}
\usepackage{booktabs}

\newcommand{\minisection}[1]{\vspace{.01in}\noindent{\textbf{#1}.}}

\usepackage[export]{adjustbox}
\usepackage{multirow}

\usepackage{algorithmic}

\DeclareMathOperator*{\argmin}{argmin}

\usepackage{times}
\usepackage{latexsym}

\usepackage[T1]{fontenc}

\usepackage[utf8]{inputenc}

\usepackage{microtype}

\usepackage{inconsolata}

\usepackage{graphicx}

\title{Analytical FFN-to-MoE Restructuring via Activation Pattern Analysis}

\author{Zehua Pei$^1$, 
Hui-Ling Zhen$^2$, 
Lancheng Zou$^1$, 
Xianzhi Yu$^2$, 
Wulong Liu$^2$, \\
\bf Sinno Jialin Pan$^1$,
Mingxuan Yuan$^2$, 
Bei Yu$^1$\\
$^1$The Chinese University of Hong Kong\quad
$^2$Huawei Technologies Co., Ltd
}

\begin{document}
\maketitle

\begin{abstract}

Scaling large language models (LLMs) improves performance but significantly increases inference costs, with feed-forward networks (FFNs) consuming the majority of computational resources. 
While Mixture-of-Experts (MoE) architectures can reduce this cost through sparse activation, restructuring existing dense models into MoEs typically requires extensive retraining on hundreds of billions of tokens.
We propose an analytical post-training framework that rapidly restructures FFNs into sparse MoE architectures using only a small calibration dataset. 
The method analyzes neuron activation patterns to partition neurons into always-active shared experts and conditionally activated routed experts, then constructs a router analytically from representative neuron statistics, enabling immediate deployment or optional lightweight fine-tuning. 
This approach applies both to dense models and recursively to existing MoE models for hierarchical sparsity.
Experiments demonstrate up to $1.17\times$ speedup in compute-bound scenarios with only minutes of processing and 2k-sample fine-tuning, outperforming methods requiring orders of magnitude more resources.
\footnote{Code: \url{https://github.com/JarvisPei/CMoE}}

\end{abstract}

\section{Introduction}

Large language models (LLMs) have achieved strong performance on a wide range of tasks \cite{zhang2022opt, touvron2023llama, liu2024visual, liu2024deepseek}, but their ever-growing size presents deployment challenges due to high computational demands, especially on resource-constrained hardware or under strict latency budgets. 
This has spurred the development of various inference acceleration techniques. 
Among these, the mixture-of-experts (MoE) architecture \cite{lepikhin2020gshard, du2022glam, fedus2022switch, dai2024deepseekmoe} decouples model capacity from computational cost by using a router to dynamically select a sparse subset of parameters for each input token. 
However, reaping the benefits of MoE models has traditionally required expensive pre-training from scratch, establishing a challenging trade-off between model performance and training cost.

The computational bottleneck in modern transformer architectures is disproportionately located in the feed-forward network (FFN) blocks. 
Several studies have reported high activation sparsity in FFN neurons \cite{dejavu, moefication, pei2024fusegpt}, meaning only a small fraction of neurons activate for any given input. 
This natural sparsity presents an opportunity to accelerate inference without the cost of pre-training. 
Prior work has explored restructuring dense models into MoEs, but existing methods either rely on weight-based clustering~\cite{moefication,qiu2023unlocking} that treats all neurons uniformly regardless of their activation behavior, or require extensive continual training~\cite{llama-moe,llama-moe-v2} with up to 200 billion tokens to recover quality. 
A key observation overlooked by these methods is that neuron activation frequencies exhibit two distinct groups: the majority of neurons activate only for specific inputs, while a subset activates consistently across inputs. 
By treating all neurons uniformly, prior approaches scatter consistently active neurons across experts, requiring extensive training to learn effective routing. 
This motivates an analytical approach that explicitly leverages this distinct structure: grouping high-frequency neurons into shared experts and low-frequency neurons into routed experts, with routing derived directly from activation statistics.

To overcome these limitations, we propose an analytical post-training framework that improves the performance-cost trade-off for LLM acceleration. 
The framework restructures FFNs through a rapid, analytical process using only a tiny calibration dataset. 
It operates by analyzing neuron activation patterns to distinguish frequently active neurons (grouped into `shared' experts) from sparsely active ones. 
The sparsely active neurons are then clustered into specialized `routed' experts using a balanced assignment algorithm \cite{jonker1988shortest}. 
This restructuring is broadly applicable: it can transform a dense model's single, large FFN into a sparse MoE architecture, or it can be applied recursively to the individual experts of an existing MoE model to induce a finer-grained hierarchical sparsity. 
The framework constructs a router analytically from activation statistics, bypassing the need for expensive router training and enabling rapid deployment with a training-free baseline or optional lightweight fine-tuning.

Our contributions are:

\begin{itemize}
    \item We identify two distinct groups in FFN neuron activation frequencies: a subset activates consistently across inputs while the majority activates conditionally. We show that prior FFN-to-MoE methods overlook this structure.
    \item We propose an analytical framework that leverages this observation to partition neurons into shared and routed experts, constructing a router directly from representative neuron statistics without training.
    \item The method outperforms prior MoE restructuring methods in accuracy with only 2k-sample fine-tuning, and achieves up to $1.17\times$ inference speedup. It applies to both dense models and existing MoE architectures for hierarchical sparsity.
\end{itemize}

\section{Related Work}

In contrast to pretraining MoE models from scratch, recent research has investigated constructing MoE architectures by repurposing existing dense LLMs. 
Current methodologies generally follow two paradigms: (1) partitioning FFN parameters while preserving total parameter count~\cite{zuo2022moebert,moefication,yang2024xmoe}, or (2) expanding capacity while retaining activation dimensions~\cite{komatsuzaki2022sparse,wu2024parameter}. 
This work prioritizes the former.
MoEBERT~\cite{zuo2022moebert} redistributes top-scoring neurons across experts using an importance-driven strategy; while it also shares important neurons, it duplicates them inside every expert rather than forming structurally separate shared experts, and relies on task-specific gradient-based importance scores rather than task-agnostic activation profiling.
MoEfication~\cite{moefication} leverages sparse activation patterns in ReLU-based FFNs, decomposing layers into expert groups with a learned router.
G-MoEfication~\cite{lee2024breaking} generalizes MoEfication to non-ReLU models by retaining representative values for unselected experts.
LLaMA-MoE~\cite{llama-moe} and its successor~\cite{llama-moe-v2} partition FFNs into experts but require extensive continual training (200B and 7B tokens, respectively).
EMoE~\cite{qiu2023unlocking} clusters neurons by key vectors during fine-tuning, while Read-ME~\cite{cai2024textit} focuses on domain-aware expert construction with system co-design. 
In contrast, our method analytically restructures FFNs by partitioning neurons into shared and routed experts based on activation patterns, then constructs a router from representative neuron statistics, requiring only 2k-sample fine-tuning.

A parallel line of work studies fully differentiable routing. 
ReMoE~\cite{wang2024remoe} replaces hard Top-K with ReLU routing, and Lory~\cite{zhong2024lory} performs differentiable expert merging at large token budgets.
These methods learn routers during pre-training, whereas our training-light analytical restructuring is complementary.

Orthogonal to FFN-to-MoE conversion, other efficiency techniques include structured pruning methods such as SliceGPT~\cite{ashkboos2024slicegpt} and SLEB~\cite{song2024sleb}, which remove model components statically.
Activation sparsity methods exploit natural sparsity in FFN hidden states: DejaVu~\cite{dejavu} leverages contextual sparsity, while TEAL~\cite{liu2024training} and WINA~\cite{chen2025wina} use magnitude- or weight-informed thresholds.
Learn-To-be-Efficient~\cite{zheng2024learn} trains models to activate fewer neurons.
These approaches operate at different granularities and can be complementary to our MoE restructuring.

\section{Motivation: Activation Patterns}
\label{sec:motivation}

Before presenting our method, we analyze the activation patterns in FFN layers. These observations inform the design of our analytical restructuring approach.

\subsection{High Activation Sparsity}
\label{sec:obs1}

Consider an FFN layer that takes input $\mathbf{x} \in \mathbb{R}^d$ and produces output $F(\mathbf{x}) \in \mathbb{R}^d$. 
The computation involves a hidden state $\mathbf{h} \in \mathbb{R}^{d_h}$, where $d_h$ is the hidden dimension (typically $d_h \gg d$).
Each neuron's contribution to the output can be written as:
\begin{equation}
\label{eq:decompose}
    F(\mathbf{x}) = \sum_{i=1}^{d_h} h_i \mathbf{w}_i,
\end{equation}
where $h_i$ is the $i$-th hidden activation and $\mathbf{w}_i \in \mathbb{R}^d$ is the corresponding column of the output projection.

\begin{figure}[tb!]
    \centering
    \includegraphics[width=0.85\linewidth]{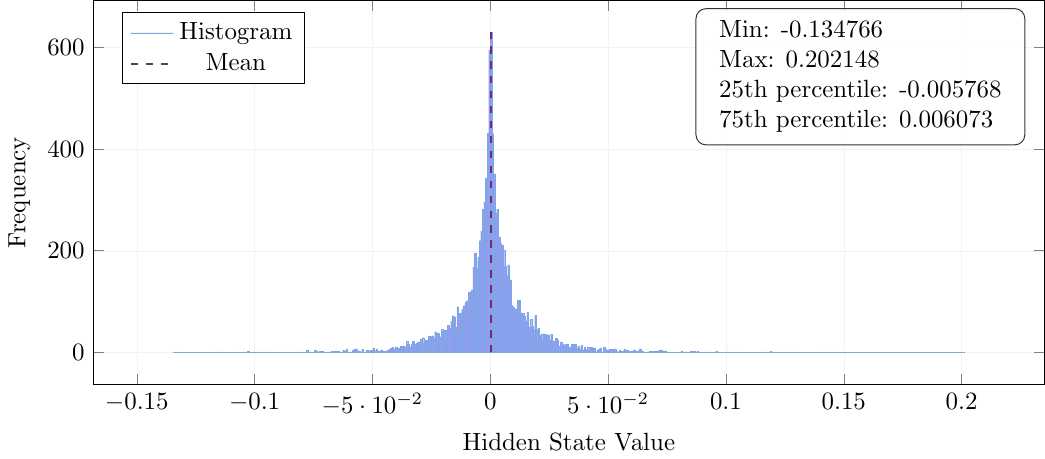}
    \caption{Distribution of FFN hidden state $\mathbf{h}$ (Llama-2-7B). Most activations concentrate near zero.}
    \label{fig:hs_motivation}
\end{figure}

\Cref{fig:hs_motivation} shows the distribution of hidden activations for a representative layer. The distribution is sharply peaked at zero, indicating that most neurons contribute negligibly to the output for any given input. This sparsity suggests that selectively activating neurons could reduce computation while preserving output quality. However, to exploit this, we need to understand \textit{which} neurons are important for \textit{which} inputs.

\subsection{Diverse Activation Patterns}
\label{sec:obs2}

To characterize neuron behavior across inputs, we compute activation rates over a calibration set. For each neuron $i$, we define its activation rate $\mu_i$ as the fraction of tokens for which it ranks among the top-$K_a$ activations by magnitude, where $K_a$ is a hyperparameter specifying how many top activations to consider.

\begin{figure}[tb!]
    \centering
    \includegraphics[width=0.85\linewidth]{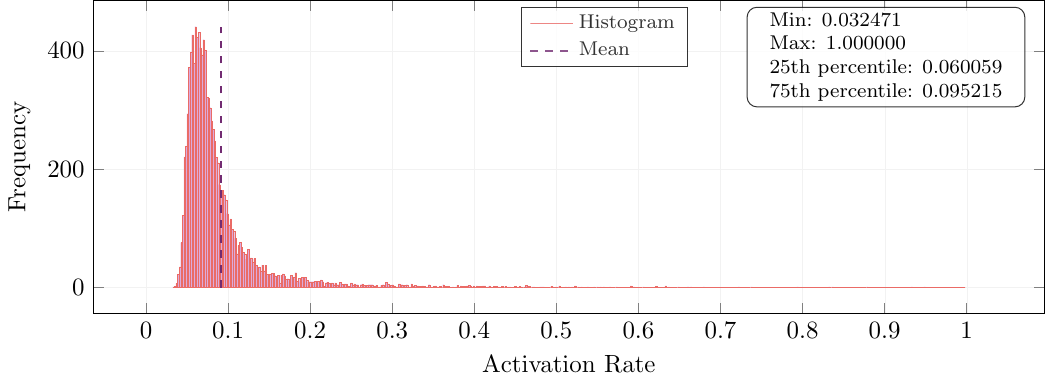}
    \caption{Distribution of neuron activation rates $\boldsymbol{\mu}$ (Llama-2-7B, $K_a=1000$ for visualization; similar patterns hold at smaller $K_a$). Most neurons have low activation rates, while a subset is consistently active.}
    \label{fig:rates_motivation}
\end{figure}

\Cref{fig:rates_motivation} reveals two distinct groups. The majority of neurons exhibit low activation rates (peak near 0.07), meaning they activate only for specific inputs. However, a subset of neurons shows consistently high activation rates approaching 1, indicating they contribute to nearly all inputs. This distinction suggests a natural partition: neurons that are always important versus neurons that are conditionally important.

This observation has implications for MoE design. Prior methods~\cite{moefication,qiu2023unlocking} treat all neurons uniformly when constructing experts, ignoring this bimodal pattern. Since these methods do not distinguish consistently active neurons from conditionally active ones, high-frequency neurons may be scattered across different routed experts. This forces the router to activate most experts regardless of input, undermining the sparsity that MoE relies on for efficiency. In contrast, explicitly separating high-frequency neurons into always-active shared experts and grouping low-frequency neurons into conditionally activated routed experts yields a more structured architecture where the router only needs to select among genuinely input-dependent experts.

\subsection{Problem Formulation}
\label{sec:problem}

Based on these observations, we formulate the restructuring objective. Given the original FFN output $F(\mathbf{x})$, we seek an MoE architecture that minimizes reconstruction error:
\begin{equation}
\label{eq:objective}
    \min \mathbb{E}_{\mathbf{x}} \left[ \| F_{MoE}(\mathbf{x}) - F(\mathbf{x}) \|^2 \right].
\end{equation}
The MoE output is defined as $F_{MoE}(\mathbf{x}) = E^s(\mathbf{x}) + \sum_{i=1}^{N_r} g_i \cdot E_i^r(\mathbf{x})$, where $E^s$ is a shared expert that is always active, $\{E_i^r\}_{i=1}^{N_r}$ are $N_r$ routed experts, and $g_i \in \{0,1\}$ are gate values produced by a router.

Since experts are constructed by partitioning the original neurons without adding parameters, the objective reduces to determining which routed experts to deactivate. When some routed experts are deactivated (receiving $g_i = 0$), the reconstruction error equals the sum of their outputs. To minimize this error, we should construct the expert partition and router such that deactivated experts have minimal expected contribution. 

The observations above suggest that shared experts should contain high-frequency neurons, routed experts should group co-activated neurons, and the router can be derived from activation patterns. The following section presents our analytical solution.

\section{Methodology}
\label{sec:method}

\begin{figure}[tb!]
    \centering
    \includegraphics[width=0.95\linewidth]{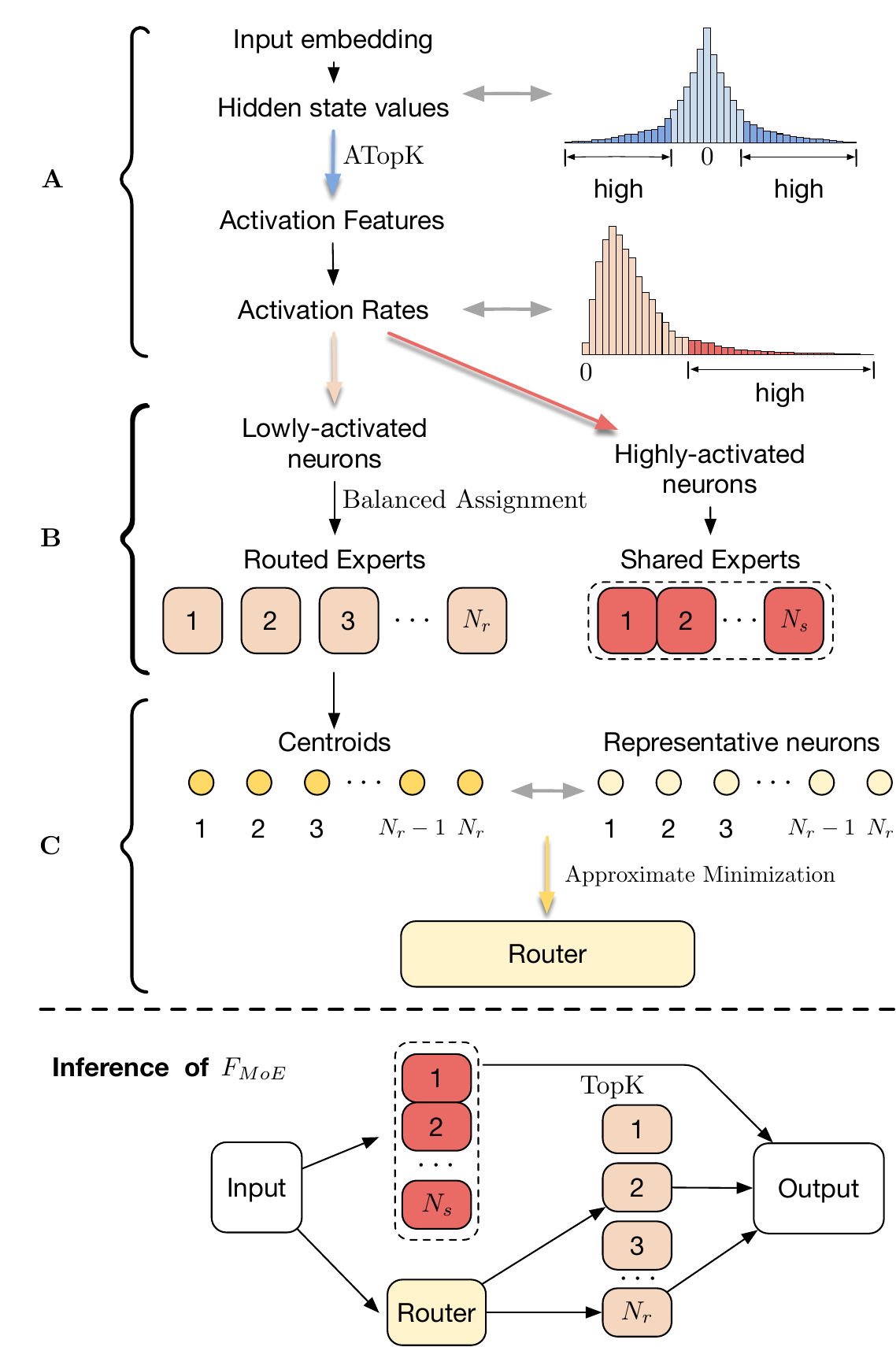}
    \caption{Overview of the proposed analytical FFN-to-MoE restructuring framework.}
    \label{fig:cmoe}
\end{figure}

Building on the problem formulation in \Cref{sec:problem}, we now present our analytical solution. The framework operates in three stages: (A) profiling neuron activation patterns, (B) partitioning neurons into shared and routed experts, and (C) constructing an analytical router. \Cref{fig:cmoe} illustrates the overall pipeline.

\subsection{Expert Construction}
\label{sec:expert_cons}

For an FFN with hidden dimension $d_h$, we construct $N$ experts of size $m = d_h / N$, comprising $N_s$ shared experts and $N_r$ routed experts ($N_s + N_r = N$). For LLaMA-style models with SwiGLU activation, the FFN computes:
\begin{align}\label{eq:ori_ffn_method}
\mathbf{h} &= \text{Swish}(\mathbf{W}_{\text{gate}}^\top \mathbf{x}) \odot (\mathbf{W}_{\text{up}}^\top \mathbf{x}), \nonumber \\
F(\mathbf{x}) &= \mathbf{W}_{\text{down}}^\top \mathbf{h},
\end{align}
with $\mathbf{x} \in \mathbb{R}^{d}$, $\mathbf{W}_{\text{up}}, \mathbf{W}_{\text{gate}} \in \mathbb{R}^{d \times d_h}$, and $\mathbf{W}_{\text{down}} \in \mathbb{R}^{d_h \times d}$.

\minisection{Activation Profiling}
Using a small calibration dataset, we compute hidden states $\mathbf{H} \in \mathbb{R}^{q \times d_h}$ for $q$ tokens. For each token, we identify the top-$K_a$ neurons by activation magnitude, i.e., an absolute top-$K$ (ATopK) selection over $|h_i|$, yielding a binary activation matrix $\mathbf{A} \in \{0,1\}^{q \times d_h} = [\mathbf{c}_1\; \mathbf{c}_2 \; \cdots \; \mathbf{c}_{d_h}]$. Each column $\mathbf{c}_i$ represents neuron $i$'s activation pattern across the calibration set, and the activation rate $\mu_i = \text{mean}(\mathbf{c}_i)$ measures how frequently neuron $i$ is active (see \Cref{appendix:pipeline} for the complete pipeline).

\minisection{Shared Experts}
Based on the diverse activation pattern observed in \Cref{sec:obs2}, we select the $N_s \cdot m$ neurons with highest activation rates to form the shared expert $E^s$. These neurons are consistently active across inputs and capture common knowledge. The shared expert weights are constructed by slicing the original FFN matrices according to the selected neuron indices.

\minisection{Routed Experts}
The remaining neurons are partitioned into $N_r$ routed experts. Since neurons with similar functions tend to co-activate, we cluster them based on the similarity of their activation feature vectors $\mathbf{c}_i$. We employ a balanced assignment algorithm that groups neurons into $N_r$ equal-sized clusters while minimizing intra-cluster distance. Details are provided in \Cref{appendix:clustering}.

The resulting MoE architecture computes:
\begin{equation}
\label{eq:moe_output}
F_{MoE}(\mathbf{x}) = E^s(\mathbf{x}) + \sum_{i=1}^{N_r} g_i \cdot E_i^{r}(\mathbf{x}),
\end{equation}
where $g_i \in \{0,1\}$ are gate values and only the top-$N_k$ routed experts (by router score) are activated, with $N_k$ denoting the number of routed experts selected per input.

\subsection{Analytical Router Construction}
\label{sec:router_cons}

From the problem formulation in \Cref{sec:problem}, minimizing reconstruction error requires that deactivated experts contribute minimally to the output. Let $S_{de}$ denote the set of deactivated routed experts. The reconstruction error is:
\begin{equation}
\label{eq:recon_error}
F_{MoE}(\mathbf{x}) - F(\mathbf{x}) = -\sum_{i \in S_{de}} E_i^r(\mathbf{x}).
\end{equation}

Each expert's output can be decomposed as a sum over its neurons' contributions. Under the sparsity observation from \Cref{sec:obs1}, neurons with small hidden activations contribute negligibly (see \Cref{appendix:sparsity} for the formal hypothesis). This suggests that the $L_1$ magnitude of an expert's hidden state $\|\mathbf{h}_i^r\|_1$ serves as a proxy for its output magnitude. Thus, minimizing reconstruction error approximately reduces to (full derivation in \Cref{appendix:router}):
\begin{equation}
\label{eq:reduced_objective}
\min_{G} \mathbb{E}_{\mathbf{x}}\left[\sum_{i \in S_{de}} \|\mathbf{h}_i^r\|_1\right],
\end{equation}
where $G$ is the router that determines $S_{de}$ via top-$N_k$ selection.

This objective is minimized when the router scores $\mathbf{s} = [s_1, \ldots, s_{N_r}]$ rank experts by their expected hidden state magnitude, ensuring that experts with larger contributions are activated while those with smaller contributions are deactivated.

\minisection{Representative Neuron Selection}
To construct such a router analytically, we identify a \textbf{representative neuron} $R_j$ for each expert $j$ as the neuron whose activation pattern is closest to the cluster centroid $\hat{\mathbf{c}}_j$:
\begin{equation}
\label{eq:rep_neuron}
R_j = \argmin_{i \in \text{cluster } j} \|\mathbf{c}_i - \hat{\mathbf{c}}_j\|_2,
\end{equation}
where $\hat{\mathbf{c}}_j$ is the centroid of cluster $j$ obtained from the balanced clustering step. Since $R_j$ best represents the expert's typical activation behavior, its hidden activation $h_{R_j}$ approximates the expert's overall contribution.

The router is then constructed using only the representative neurons' parameters:
\begin{equation}
\label{eq:router}
G(\mathbf{x}) = \text{Swish}(\mathbf{W}_{\text{gate}}^{R\top} \mathbf{x}) \odot (\mathbf{W}_{\text{up}}^{R\top} \mathbf{x}),
\end{equation}
where $\mathbf{W}_{\text{gate}}^{R}, \mathbf{W}_{\text{up}}^{R} \in \mathbb{R}^{d \times N_r}$ contain only the columns corresponding to representative neurons. This yields router scores $\mathbf{s} = [s_1, \ldots, s_{N_r}] = G(\mathbf{x})$ that approximate each expert's expected contribution, enabling effective routing.

\subsection{Fine-tuning Enhancements}
\label{sec:finetuning}

The analytical router provides a training-free baseline. For further improvement, we introduce two enhancements for optional lightweight fine-tuning.

\minisection{Learnable Scaling}
The initial gate values are binary ($g_i \in \{0,1\}$). To enable gradient-based optimization, we introduce learnable scaling parameters $\mathbf{u} = [u_1, \ldots, u_{N_r}]$, initialized to zero. For selected experts, the gate becomes $g_i = 1 + s'_i \cdot u_i$, where $\mathbf{s}' = \text{Softmax}(\mathbf{s})$.

\minisection{Load Balancing}
To ensure balanced expert utilization without auxiliary losses, we introduce adaptive bias terms $\mathbf{b} = [b_1, \ldots, b_{N_r}]$ added to scores before top-$N_k$ selection~\cite{liu2024deepseek}. The final gating logic is:
\begin{equation}
\label{eq:final_gate}
g_i = \begin{cases}
1 + s'_i \cdot u_i, & \text{if } s'_i + b_i \in \text{Top-}N_k, \\
0, & \text{otherwise}.
\end{cases}
\end{equation}
The biases are updated after each step: if expert $i$ is overloaded ($p_i > p^*$), we decrease $b_i$ by $\gamma$; if underloaded ($p_i < p^*$), we increase $b_i$ by $\gamma$, where $p_i$ is expert $i$'s utilization fraction, $p^* = 1/N_r$ is the uniform target, and $\gamma = 10^{-3}$. 

\subsection{Application to Existing MoE Models}
\label{sec:moe_application}

The framework applies not only to dense FFNs but also to existing MoE models. For an MoE layer with experts $\{E_i\}$, we apply our restructuring to each expert individually, transforming it into a hierarchical structure with shared and routed sub-experts:
\begin{equation}
E_i(\mathbf{x}) \rightarrow E^s_i(\mathbf{x}) + \sum_{j=1}^{N'_r} g'_{i,j} \cdot E^r_{i,j}(\mathbf{x}).
\end{equation}
This creates a two-level hierarchy: the top-level router selects primary experts, and within each activated expert, a sub-router selects specialized sub-experts. This induces finer-grained sparsity for further acceleration.

\section{Experiments}
\label{sec:exp}

We evaluate the proposed framework as a post-training sparsification method for inference acceleration on large language models.

\subsection{Experimental Settings}
\label{sec:exp_settings}

\minisection{Models and Implementation}
We evaluate on Llama-2 7B, Llama-2 70B, Qwen-2.5-7B, and Qwen-3-30B-A3B using Hugging Face Transformers~\cite{wolf2019huggingface} and PyTorch~\cite{paszke2019pytorch}. Qwen-2.5 72B is additionally used in \Cref{tab:industrial_speedup} for the industrial speedup evaluation.

\minisection{Calibration}
We use 8 examples (2048 tokens each) from WikiText-2~\cite{merity2016pointer} for activation profiling, with $K_a=10$ for top-$K$ activation selection.

\minisection{Fine-tuning}
We apply LoRA~\cite{hu2021lora} (rank 8, alpha 32) on 2,048 WikiText-2 samples for 1 epoch using Adam~\cite{kingma2014adam} ($\beta_1=0.9$, $\beta_2=0.95$), with learning rates 0.001 for router scaling and 5.95e-5 for LoRA. Load balancing uses $\gamma=0.001$.

\minisection{Baselines}
We compare against: (1) \textit{Structured Pruning}: SliceGPT~\cite{ashkboos2024slicegpt} and SLEB~\cite{song2024sleb} at 20\% reduction; (2) \textit{MoE Restructuring}: LLaMA-MoE~\cite{llama-moe}, LLaMA-MoE-v2~\cite{llama-moe-v2}, and EMoE~\cite{qiu2023unlocking}. All baselines are re-implemented by us and fine-tuned with LoRA under matched data budget (2k samples) for fair comparison. Structured pruning baselines use 20\% reduction because they reduce both FFN and attention parameters, making their effective FFN sparsity comparable to our 25\% FFN-only sparsity.

\minisection{Configuration}
Unless otherwise noted, we use 25\% sparsity (i.e., 75\% of FFN neurons are activated per token) with S3A3E8 (3 shared + 3 active routed / 8 total experts). All MoE methods use 8 experts for fair comparison.

\subsection{Main Results}
\label{sec:main_results}

\begin{table}[tb!]
\caption{Zero-shot accuracy (\%) at 25\% sparsity (Sp.). All sparsified methods are fine-tuned with 2k samples; the dense baseline is not fine-tuned.}
\label{table:zero_shot}
\centering
\resizebox{\columnwidth}{!}{
\begin{tabular}{l|c|c c c c c}
\toprule
Method & Sp. & PIQA & WinoG. & ARC-E & ARC-C & HellaS.  \\
\midrule
\multicolumn{7}{c}{\textbf{Llama-2 7B}} \\
\midrule
Dense & 0\% & 78.78	& 69.06	& 74.58	& 46.16	& 76.00 	\\
\midrule
SliceGPT & 20\% & 65.71	& 62.88	& 59.76	& 33.21	& 51.34  	\\
SLEB & 20\% & 73.13	& 58.98	& 57.90	& 33.02	& 62.47  	\\
\midrule
LLaMA-MoE & 25\% & 49.35	& 50.28	& 54.04	& 26.37	& 25.77  	\\
LLaMA-MoE-v2 & 25\% & 63.55	& 59.35	& 63.77	& 34.81	& 54.89  	\\
EMoE & 25\% & 72.47	& 64.48	& 58.63	& 35.75	& 60.80  	\\
\midrule
\textbf{Ours} & 25\% & \textbf{74.34}	&\textbf{65.77}	&\textbf{67.09}		&\textbf{40.35}	&\textbf{69.36} \\
\midrule
\multicolumn{7}{c}{\textbf{Llama-2 70B}} \\
\midrule
Dense & 0\% & 82.70	& 77.98	& 80.98		& 57.34	& 83.84  	\\
\midrule
SliceGPT & 20\% & 68.91	& 70.06	& 64.56	& 41.14	& 56.26  	\\
SLEB & 20\% & 77.39	& 65.55	& 62.37	& 40.11	& 68.39  	\\
\midrule
LLaMA-MoE & 25\% & 51.95	& 56.50	& 59.09	& 32.40	& 27.57  	\\
LLaMA-MoE-v2 & 25\% & 66.79	& 66.57	& 68.94	& 42.38	& 59.57  	\\
EMoE & 25\% & 76.34	& 72.33	& 63.47	& 43.62	& 66.19  	\\
\midrule
\textbf{Ours} & 25\% & \textbf{78.49}	&\textbf{73.49}	&\textbf{73.32}		&\textbf{49.86}	&\textbf{76.12}  \\
\midrule
\multicolumn{7}{c}{\textbf{Qwen-2.5-7B}} \\
\midrule
Dense & 0\% & 79.82	& 73.16	& 77.36	& 51.02	& 78.86  	\\
\midrule
SliceGPT & 20\% & 66.19	& 66.51	& 61.88	& 36.69	& 53.21  	\\
SLEB & 20\% & 74.95	& 61.76	& 59.95	& 35.80	& 64.41  	\\
\midrule
LLaMA-MoE & 25\% & 49.63	& 53.21	& 57.05	& 28.64	& 25.65  	\\
LLaMA-MoE-v2 & 25\% & 64.25	& 62.71	& 65.77	& 37.59	& 56.06  	\\
EMoE & 25\% & 73.98	& 65.41	& 60.63	& 38.48	& 62.71  	\\
\midrule
\textbf{Ours} & 25\% & \textbf{75.93}	&\textbf{69.36}	&\textbf{70.59}		&\textbf{43.86}	&\textbf{72.21}  \\
\midrule
\multicolumn{7}{c}{\textbf{Qwen-3-30B-A3B}} \\
\midrule
Dense & 0\% & 84.51	& 79.18	& 84.43	& 57.88	& 87.44  	\\
\midrule
SliceGPT & 20\% & 70.60	& 71.58	& 66.88	& 41.85	& 58.41  	\\
SLEB & 20\% & 79.16	& 66.01	& 70.08	& 42.11	& 71.74  	\\
\midrule
LLaMA-MoE & 25\% & 52.18	& 54.48	& 62.50	& 30.77	& 28.32  	\\
LLaMA-MoE-v2 & 25\% & 65.54	& 67.24	& 71.27	& 41.99	& 62.78  	\\
EMoE & 25\% & 74.76	& 70.50	& 65.78	& 43.12	& 70.62  	\\
\midrule
\textbf{Ours} & 25\% & \textbf{80.23}	&\textbf{74.84}	&\textbf{76.75}		&\textbf{48.80}	&\textbf{80.71}  \\
\bottomrule
\end{tabular}
}
\end{table}

\minisection{Zero-Shot Downstream Tasks}
\Cref{table:zero_shot} presents results on five benchmarks: PIQA~\cite{bisk2020piqa}, WinoGrande~\cite{sakaguchi2021winogrande}, ARC-Easy, ARC-Challenge~\cite{clark2018think}, and HellaSwag~\cite{zellers2019hellaswag}.
At 25\% sparsity, our method consistently outperforms all baselines across four models spanning 7B to 30B parameters.
On Llama-2 7B, it achieves 74.34\% PIQA and 69.36\% HellaSwag, exceeding both structured pruning and MoE restructuring approaches.
The improvements scale consistently: on Llama-2 70B we observe similar gains, and on the Qwen-3-30B-A3B, our method achieves 80.23\% PIQA and 80.71\% HellaSwag.

\minisection{Broader Evaluation on Knowledge, Coding, and Math}
Beyond these five zero-shot tasks, we also evaluate Llama-2 7B at 25\% sparsity (S3A3E8) on MMLU-5shot, HumanEval pass@1, and GSM8K-8shot to cover knowledge-intensive and reasoning benchmarks. As summarized in \Cref{tab:llama2_downstream_breadth}, our analytical MoE restructuring achieves 44.02\% MMLU-5shot and competitive coding/math accuracy, outperforming LLaMA-MoE variants and EMoE across all three benchmarks. 
\begin{table}[tb!]
    \centering
    \caption{Broader evaluation on Llama-2 7B at 25\% sparsity (S3A3E8).}
    \label{tab:llama2_downstream_breadth}
    \resizebox{0.8\columnwidth}{!}{
    \begin{tabular}{l|c|c|c}
        \toprule
        \multirow{2}{*}{Method} & MMLU & HumanEval & GSM8K \\
        & (\%) & (\%) & (\%) \\
        \midrule
        LLaMA-MoE & 35.09 & 7.58 & 7.41 \\
        LLaMA-MoE-v2 & 38.02 & 9.32 & 10.09 \\
        EMoE & 43.11 & 10.29 & 12.55 \\
        \textbf{Ours} & \textbf{44.02} & \textbf{11.22} & \textbf{13.01} \\
        \bottomrule
    \end{tabular}
    }
\end{table}

\subsection{Ablation Studies}
\label{sec:ablations} 

\begin{figure}[tb!]
    \centering
    \includegraphics[width=\linewidth]{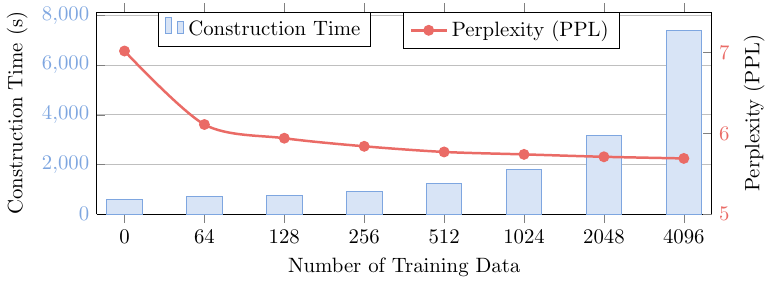}
    \caption{Data efficiency: performance and construction time vs fine-tuning samples (25\% sparsity).}
    \label{fig:app_ab_1}
\end{figure}

\minisection{Training-Free vs Fine-Tuned Performance}
\Cref{fig:app_ab_1} shows data efficiency at 25\% sparsity. The method achieves reasonable performance immediately after construction with zero fine-tuning, indicating that the analytical router initialization is effective. Performance plateaus quickly with additional data: as few as 1,024 samples reach near-optimal results.

\Cref{tab:training_free_vs_ft} compares with LLaMA-MoE-v2 on Llama-2 7B. Our training-free model achieves 42.50\% MMLU-5shot (vs 38.02\% for LLaMA-MoE-v2 after fine-tuning). With 2k samples, we reach 44.02\%, indicating most gains come from analytical restructuring rather than fine-tuning.

\begin{table}[tb!]
    \centering
    \caption{Training-free vs fine-tuned on Llama-2 7B (25\% sparsity).}
    \label{tab:training_free_vs_ft}
    \resizebox{0.95\columnwidth}{!}{
    \begin{tabular}{l|l|c|c|c}
        \toprule
        \multirow{2}{*}{Method} & \multirow{2}{*}{Regime} & MMLU & PPL & PPL \\
        & & (\%) & Wiki & C4 \\
        \midrule
        LLaMA-MoE-v2 & Training-free & 30.33 & $>$10k & $>$7k \\
        LLaMA-MoE-v2 & Fine-tuning & 38.02 & 8.68 & 19.76 \\
        \midrule
        Ours & Training-free & 42.50 & 7.32 & 11.98 \\
        Ours & Fine-tuning (2k) & \textbf{44.02} & \textbf{5.92} & \textbf{11.21} \\
        \bottomrule
    \end{tabular}
    }
\end{table}


\minisection{Calibration Sensitivity}
\Cref{tab:calibration_sensitivity} varies calibration source (WikiText-2 and C4) and size on Llama-2 7B. For WikiText-2, increasing from 8 to 64 samples yields modest MMLU gains (44.02$\rightarrow$44.89) and small perplexity reductions. Perplexity is consistent across both calibration sources, confirming the diverse activation pattern (\Cref{sec:obs2}) is intrinsic to pre-trained FFNs rather than data-specific. The low sample requirement (8 examples) makes domain-specific calibration practical when needed.

To further verify domain invariance, we profile shared-expert neurons separately on math and science data (Nemotron-Post-Training-Dataset-v1~\cite{NemotronPostTrainingDatasetV1}) and code data (OpenCoder~\cite{Huang2024OpenCoderTO}), then measure pairwise overlap. The overlap is 84\% (math/science), 86\% (math/code), and 80\% (science/code), confirming that shared-expert selection reflects intrinsic model structure rather than domain-specific artifacts.

\begin{table}[tb!]
    \centering
    \caption{Calibration sensitivity on Llama-2 7B (25\% sparsity). Perplexity is consistent across calibration sources and sizes.}
    \label{tab:calibration_sensitivity}
    \resizebox{0.75\columnwidth}{!}{
    \begin{tabular}{l|c|c|c|c}
        \toprule
        \multirow{2}{*}{Source} & \multirow{2}{*}{$n$} & MMLU & PPL & PPL \\
        & & (\%) & Wiki & C4 \\
        \midrule
        WikiText-2 & 8  & 44.02 & 5.92 & 11.21 \\
        WikiText-2 & 32 & 44.63 & 5.72 & 11.15 \\
        WikiText-2 & 64 & \textbf{44.89} & \textbf{5.69} & \textbf{10.98} \\
        \midrule
        C4         & 8  & 42.31 & 7.04 & 9.17 \\
        C4         & 32 & 43.25 & 6.92 & 9.07 \\
        C4         & 64 & \textbf{43.39} & \textbf{6.78} & \textbf{9.02} \\
        \bottomrule
    \end{tabular}
    }
\end{table}

\minisection{Clustering and Routing}
To isolate the contributions of clustering and routing, \Cref{tab:mmlu_moe} compares methods under identical settings (25\% sparsity, 2k fine-tuning). MoEfication~\cite{moefication}, G-MoEfication~\cite{lee2024breaking} (its non-ReLU generalization), and Read-ME~\cite{cai2024textit} achieve only 35.17\%, 36.37\%, and 31.24\% MMLU respectively, while our method reaches 44.02\%. Replacing each method's router with our analytical router improves results by +2 to 6 pp, and further switching to our activation-based clustering with shared experts yields an additional +5 to 7 pp, confirming both components contribute independently.

Weight-based methods assume neurons with similar parameters serve similar functions, but ignore input-dependent behavior. Our approach groups neurons that co-activate, aligning expert structure with actual usage patterns.
Notably, MoEfication also explored co-activation graph clustering, but their parameter-based method performed better. Our approach differs: (1) we explicitly separate high-frequency neurons into shared experts, and (2) we derive the router analytically from representative neurons. This combination yields nearly +9 pp over MoEfication's best method.

\begin{table}[tb!]
    \centering
    \caption{Ablation: clustering and routing on Llama-2 7B (MMLU). All use 25\% sparsity and 2k fine-tuning. G-MoEfication~\cite{lee2024breaking} extends MoEfication to non-ReLU models.}
    \label{tab:mmlu_moe}
    \resizebox{\columnwidth}{!}{
    \begin{tabular}{l|l|l|c}
        \toprule
        \multirow{2}{*}{Method} & Expert & \multirow{2}{*}{Router} & MMLU \\
        & grouping & & (\%) \\
        \midrule
        MoEfication & Param. K-means & MLP & 35.17 \\
        G-MoEfication & Param. K-means & MLP & 36.37 \\
        READ-ME & Domain-aware & Global & 31.24 \\
        \midrule
        MoEfication + ours & Param. K-means & Analytical & 37.33 \\
        G-MoEfication + ours & Param. K-means & Analytical & 39.21 \\
        READ-ME + ours & Domain-aware & Analytical & 36.79 \\
        \midrule
        \textbf{Ours} & Activation + shared & Analytical & \textbf{44.02} \\
        \bottomrule
    \end{tabular}
    }
\end{table}

\begin{table}[tb!]
    \centering
    \caption{Token budget and conversion time comparison on Llama-2 7B. E2E: end-to-end time including fine-tuning; Construct: restructuring only.}
    \label{tab:efficiency_tokens_time}
    \resizebox{0.8\columnwidth}{!}{
    \begin{tabular}{l|c|c|c}
        \toprule
        \multirow{2}{*}{Method} & Token & E2E & Construct \\
        & budget & time & time \\
        \midrule
        \textbf{Ours} & \textbf{4M} & \textbf{46min} & \textbf{4.5min} \\
        LLaMA-MoE-v1 & 200B & Weeks & 6min\textsuperscript{\dag} \\
        LLaMA-MoE-v2 & 7B & Days & 8min\textsuperscript{\dag} \\
        \bottomrule
    \end{tabular}
    }
    \vspace{-0.5em}
    {\footnotesize\textsuperscript{\dag}\,Split-only; training time not included.}
\end{table}

\begin{table}[tb!]
    \centering
    \caption{Efficiency: FLOPs, MACs, and throughput.}
    \label{tab:efficiency_flops}
    \resizebox{\columnwidth}{!}{
    \begin{tabular}{l|l|c|c|c}
        \toprule
        \multirow{2}{*}{Model} & \multirow{2}{*}{Method} & FLOPs & MACs & Thru. \\
        & & (T/G) & (G) & (tok/s) \\
        \midrule
        \multirow{2}{*}{Llama-2 7B} 
        & Dense & 1.69T & 845.7 & 45.9 \\
        & Ours (25\%) & 1.41T \textcolor{gray}{\scriptsize -16.6\%} & 707.4 \textcolor{gray}{\scriptsize -16.3\%} & 52.7 \textcolor{gray}{\scriptsize +14.8\%} \\
        \midrule
        \multirow{2}{*}{Qwen3-30B-A3B} 
        & Dense & 778.7G & 389.3 & 1.19 \\
        & Ours (Hier.) & 634.9G \textcolor{gray}{\scriptsize -18.5\%} & 331.3 \textcolor{gray}{\scriptsize -14.9\%} & 1.36 \textcolor{gray}{\scriptsize +14.3\%} \\
        \bottomrule
    \end{tabular}
    }
\end{table}

\subsection{Analysis and Discussion}
\label{sec:analysis}

\minisection{Efficiency Comparison}
\Cref{tab:efficiency_tokens_time} compares token budgets and conversion times. Note that LLaMA-MoE performs continual pre-training while ours focuses on post-training restructuring; nonetheless, practitioners choosing between approaches may consider total cost. LLaMA-MoE-v1/v2 require 200B/7B tokens (weeks/days); our method uses only 4M tokens (2k samples $\times$ 2048 tokens) and completes in 46 minutes end-to-end (4.5 minutes for analytical construction).

\Cref{tab:efficiency_flops} reports FLOPs and throughput. On Llama-2 7B, 25\% sparsity reduces FLOPs by 16.6\% and increases throughput by 14.8\%. Hierarchical application to Qwen3-30B-A3B yields 18.5\% FLOPs savings and 14.3\% throughput gains.

\begin{figure}[tb!] 
    \centering
    \includegraphics[width=\linewidth]{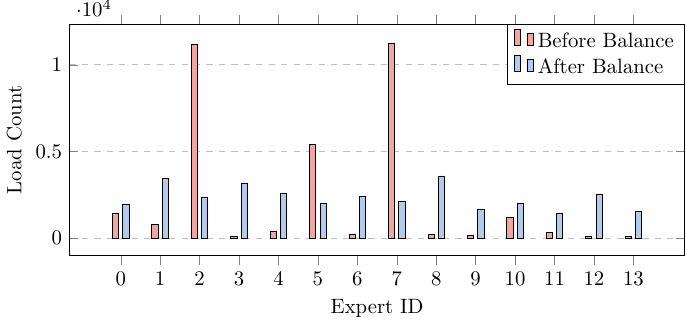}
    \caption{Load balancing: expert utilization before (left) and after (right).}
    \label{fig:ab_4}
\end{figure}

\minisection{Load Balancing}
\Cref{fig:ab_4} shows expert utilization. Without load balancing, the final layer exhibits activation skew. Our adaptive bias mechanism redistributes load uniformly, which helps achieve the full speedup potential in deployment.

\minisection{Orthogonality with Activation Sparsity}
\Cref{tab:wina_orthogonality} shows that our expert-level restructuring is orthogonal to neuron-level activation sparsity (WINA~\cite{chen2025wina}). Combining both yields 27.2\% TFLOPs reduction and 22.0\% throughput gain on Llama-2 7B, confirming they target different inefficiencies.

\begin{table}[tb!]
    \centering
    \caption{Orthogonality of neuron-level activation sparsity (WINA) and our expert-level Dense-to-MoE restructuring on Llama-2 7B at 25\% sparsity. WINA operates at the finer neuron level, while our method restructures FFNs into routed experts; combining them yields additive efficiency gains.}
    \label{tab:wina_orthogonality}
    \resizebox{\columnwidth}{!}{
    \begin{tabular}{l|c|c|c}
        \toprule
        Method & TFLOPs (↓) & GMACs (↓) & tokens/s (↑) \\
        \midrule
        Dense (baseline) & 1.69 & 845.71 & 45.88 \\
        WINA (25\% sparsity) & 1.31 (-22.5\%) & 691.19 (-18.3\%) & 51.76 (+12.8\%) \\
        Ours (25\% sparsity) & 1.41 (-16.6\%) & 707.36 (-16.3\%) & 52.67 (+14.8\%) \\
        \midrule
        \textbf{Ours + WINA} & \textbf{1.23} (-27.2\%) & \textbf{625.53} (-26.0\%) & \textbf{55.97} (+22.0\%) \\
        \bottomrule
    \end{tabular}
    }
\end{table}

\minisection{Industrial Speedup}
\Cref{tab:industrial_speedup} shows inference speedups on Qwen-2.5 72B across memory- and compute-bound (Batch Size > 400) scenarios. In compute-bound settings with 32k context, S1A5E8 configuration achieves up to $1.17\times$ speedup.

\begin{table}[tb!] 
    \centering
    \caption{Inference speedup (25\% sparsity) on Qwen-2.5 72B at different context lengths and scenarios. S$x$A$y$E$z$: $x$ shared + $y$ active routed / $z$ total experts.}
    \label{tab:industrial_speedup}
    \resizebox{0.8\columnwidth}{!}{
    \begin{tabular}{l|cc|cc}
    \toprule
    \multirow{2}{*}{Config} & \multicolumn{2}{c|}{Mem-Bound} & \multicolumn{2}{c}{Compute-Bound} \\
    & 4k ctx & 32k ctx & 4k ctx & 32k ctx \\
    \midrule
    S1A5E8 & $1.08\times$ & $1.15\times$ & $1.12\times$ & $\mathbf{1.17\times}$ \\
    S3A3E8 & $1.06\times$ & $1.13\times$ & $1.11\times$ & $1.15\times$ \\
    S2A4E8 & $1.05\times$ & $1.12\times$ & $1.10\times$ & $1.12\times$ \\
    \midrule
    S4A8E16 & $1.02\times$ & $1.10\times$ & $1.08\times$ & $1.11\times$ \\
    S6A6E16 & $1.03\times$ & $1.08\times$ & $1.07\times$ & $1.10\times$ \\
    S3A9E16 & $1.02\times$ & $1.05\times$ & $1.05\times$ & $1.09\times$ \\
    \bottomrule
    \end{tabular}
    }
    \vspace{-0.5em}
\end{table}

\begin{figure*}[tb!]
    \centering
    \includegraphics[width=\linewidth]{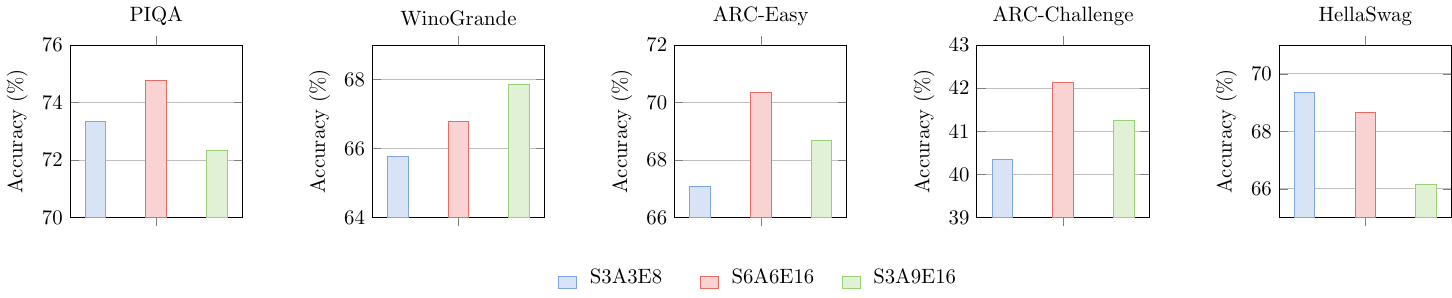}
    \caption{Impact of expert configuration at 25\% sparsity. S$x$A$y$E$z$ denotes $x$ shared experts + $y$ active routed experts out of $z$ total experts.}
    \label{fig:expert_config_comparison}
\end{figure*}

\minisection{Perplexity-Sparsity Trade-off}
\Cref{tab:sparsity_ppl} studies WikiText-2 perplexity on Llama-2 7B as we vary sparsity with 16 total experts. Interestingly, at 0.125 sparsity (87.5\% neurons active), our converted model slightly outperforms the dense baseline (5.25 vs 5.27), suggesting the restructuring may provide implicit regularization.

\begin{table}[tb!]
    \centering
    \caption{Perplexity vs sparsity on Llama-2 7B (16 experts). Lower is better.}
    \label{tab:sparsity_ppl}
    \resizebox{\columnwidth}{!}{
    \begin{tabular}{l|c|cccccc}
        \toprule
        Sparsity & Dense & 0.75 & 0.625 & 0.5 & 0.375 & 0.25 & 0.125 \\
        \midrule
        PPL $\downarrow$ & 5.27 & 12.73 & 9.56 & 7.71 & 6.55 & 5.78 & \textbf{5.25} \\
        \bottomrule
    \end{tabular}
    }
\end{table}

\minisection{Expert Configuration Impact}
\Cref{fig:expert_config_comparison} compares different expert configurations at 25\% sparsity. S6A6E16 achieves highest performance on PIQA and ARC-Easy, while S3A9E16 performs best on WinoGrande, indicating that optimal configuration depends on task characteristics.


\minisection{Self-Consistency Benefits}
\Cref{tab:self_consistency} evaluates $k$-sample self-consistency (voting) at 25\% sparsity. Interestingly, sparse models benefit more from self-consistency than dense ones: on Llama-2 7B, increasing $k$ from 1 to 5 yields +4.72 pp for our method vs only +0.57 pp for dense. On Qwen3-30B-A3B, the gap nearly closes entirely (+6.91 pp vs +0.58 pp), suggesting that variance from sparse routing can be effectively averaged out at deployment time.

\begin{table}[tb!]
    \centering
    \caption{Effect of $k$-sample self-consistency at 25\% sparsity. Accuracy (\%).}
    \label{tab:self_consistency}
    \resizebox{0.8\columnwidth}{!}{
    \begin{tabular}{l|c|c|c|c|c}
        \toprule
        Method & $k$ & PIQA & ARC-E & ARC-C & Avg \\
        \midrule
        \multicolumn{6}{c}{\textbf{Llama-2 7B}} \\
        \midrule
        Dense & 1 & 78.78 & 74.58 & 46.16 & 66.51 \\
        Dense & 5 & 79.21 & 75.29 & 46.75 & 67.08 \\
        Ours & 1 & 74.34 & 67.09 & 40.35 & 60.59 \\
        Ours & 5 & \textbf{77.52} & \textbf{73.88} & \textbf{44.54} & \textbf{65.31} \\
        \midrule
        \multicolumn{6}{c}{\textbf{Qwen3-30B-A3B}} \\
        \midrule
        Dense & 1 & 84.51 & 84.43 & 57.88 & 75.61 \\
        Dense & 5 & 85.11 & 85.33 & 58.12 & 76.19 \\
        Ours & 1 & 80.23 & 76.75 & 48.80 & 68.59 \\
        Ours & 5 & \textbf{84.56} & \textbf{84.75} & \textbf{57.19} & \textbf{75.50} \\
        \bottomrule
    \end{tabular}
    }
\end{table}

\section{Further Discussion}
\label{sec:discussion}

\minisection{Broader Impact and Future Directions}
Our work presents an analytical post-training framework for reducing the significant computational overhead of LLM inference, thereby making powerful models more accessible for research and deployment in resource-constrained settings. Beyond a pure acceleration technique, the analytical nature of the method offers a new lens for interpreting the internal workings of FFNs. The distinct grouping of neurons into `shared' and `routed' experts based on activation statistics provides empirical evidence for functional specialization within these layers. Future research could leverage this methodology to analyze how knowledge is encoded and processed within LLMs. For future work, extending this analytical restructuring approach to other parts of the transformer, such as attention heads, is a promising direction. Additionally, exploring more sophisticated analytical techniques for router construction could potentially close the remaining gap with fully trained routers, without sacrificing the efficiency of the post hoc approach.

\minisection{Compatibility with Other Efficiency Techniques}
The analytical restructuring is orthogonal to most system- and model-level efficiency methods and can be composed with them. In practice, FFN restructuring integrates well with post-training quantization (e.g., AWQ/QAT) because the operation preserves layer interfaces; it can be applied either before or after quantization with a small calibration pass to maintain accuracy. Similarly, attention-side optimizations (KV-cache compression, speculative decoding, and attention sparsity) target different bottlenecks and are complementary. 
Structured pruning (e.g., SliceGPT, SLEB) and our dynamic expert routing address different regimes: pruning induces static capacity reduction across all inputs, while our method activates capacity conditionally per token. 
Similarly, training-free activation sparsity methods (e.g., TEAL, WINA) operate at the finer neuron level and can be applied within our routed experts to further reduce FLOPs.
In deployment, load-balancing and batching policies remain important to realize end-to-end speedups; our built-in bias adaptation mitigates expert hot-spotting and improves utilization on both memory-bound and compute-bound settings. Overall, after a lightweight calibration and restructuring step, the framework serves as an FFN replacement that composes with quantization, caching, pruning, and serving optimizations to widen the practical acceleration envelope.

\section{Conclusion}
We introduced a post-training framework that analytically restructures FFNs into sparse MoE architectures using only a small calibration dataset and minutes of computation (4.5 minutes on Llama-2 7B). By leveraging the diverse distribution of neuron activation frequencies, the method partitions neurons into shared and routed experts and constructs a router from representative neuron statistics, enabling strong performance with optional lightweight fine-tuning. It applies to both dense and existing MoE models for hierarchical sparsity, and is orthogonal to techniques such as quantization and neuron-level activation sparsity, offering a practical, training-light path to deploying performant sparse LLMs.

\section{Limitations}
Our framework has three main limitations.
First, activation profiling quality depends on the calibration dataset; performance is best when the data is representative of the target domain, though the method is relatively robust to calibration set size.
Second, the discrete nature of sparse routing introduces higher generation variance, which can be mitigated via self-consistency.
Third, our evaluation focuses on English benchmarks and decoder-only transformers (Llama-2, Qwen); extension to multilingual and encoder-decoder or state-space architectures is left to future work.


\clearpage
\bibliography{custom,ref/main,ref/example_paper}

\clearpage
\appendix
\counterwithin{figure}{section}
\counterwithin{table}{section}

\section{Detailed Mathematical Derivations}
\label{appendix:1}

This appendix provides the detailed mathematical derivations and algorithmic analysis that support the core concepts presented in the main manuscript.

\subsection{Sparsity Hypothesis}
\label{appendix:sparsity}

Building on the activation sparsity observation in \Cref{sec:obs1}, we formalize the hypothesis that justifies our router construction. Given input $\mathbf{x} \in \mathbb{R}^{d}$, each neuron's contribution to the FFN output can be analyzed independently. The FFN output decomposes as:
\begin{equation}
\label{eq:sum_appendix}
    F(\mathbf{x}) = \sum_{i=1}^{d_h} h_i \mathbf{w}_{down,i}
\end{equation}
where $h_i$ is the hidden activation and $\mathbf{w}_{down,i}$ is the corresponding output projection column.

Each $h_i$ acts as a gating score for $\mathbf{w}_{down,i}$. Since structured pruning research shows that $\|F(\mathbf{x})\|$ is typically small due to residual connections, FFN activations exhibit high sparsity. This leads to our central hypothesis:
\begin{align}
\label{eq:hypo_appendix}
    \arg\min_i |h_i \mathbf{w}_{down,i}| \approx \arg\min_i |h_i|
\end{align}
This approximation is justified because when $h_i$ is extremely small, the product $h_i \mathbf{w}_{down,i}$ vanishes regardless of the magnitude of $\mathbf{w}_{down,i}$.

\subsection{Complete Activation Analysis Pipeline}
\label{appendix:pipeline}

We provide the complete mathematical pipeline for neuron activation profiling.

\textbf{Step 1: Hidden State Computation.} Given input $\mathbf{X} \in \mathbb{R}^{b\times s\times d}$, we reshape to $\mathbf{X}' \in \mathbb{R}^{q \times d}$ where $q = b \cdot s$. The hidden states are:
\begin{equation}
\label{eq:h_matrix_appendix}
    \mathbf{H} = \text{Swish}(\mathbf{X}'\mathbf{W}_{gate})\odot(\mathbf{X}'\mathbf{W}_{up})
\end{equation}
where $\mathbf{H} \in \mathbb{R}^{q \times d_h}$.

\textbf{Step 2: Activation Matrix Construction (ATopK).} For each token, we identify the top-$K_a$ neurons by activation magnitude, i.e., an absolute top-$K$ (ATopK) selection over $|h_i|$:
\begin{equation}
\label{eq:atopk_appendix}
    a_i = \begin{cases}
        1, & |h_i| \in \text{TopK}(\{|h_j| : 1 \leq j \leq d_h\}, K_a), \\
        0, & \text{otherwise},
    \end{cases}
\end{equation}
yielding a binary activation matrix $\mathbf{A} \in \{0,1\}^{q \times d_h} = [\mathbf{c}_1\; \mathbf{c}_2 \; \cdots \; \mathbf{c}_{d_h}]$, where each column $\mathbf{c}_i \in \{0,1\}^{q}$ is the activation feature vector for neuron $i$.

\textbf{Step 3: Activation Rate Computation.}
\begin{equation}\label{eq:mu_appendix}
    \mu_i = \frac{1}{q} \sum_{t=1}^{q} \mathbf{A}[t,i], \quad \boldsymbol{\mu} = [\mu_1, \cdots, \mu_{d_h}]
\end{equation}

\textbf{Step 4: Shared Expert Selection.} We select the $N_s \cdot m$ neurons with highest $\mu_i$:
\begin{equation}
    \label{eq:S_Ns_appendix}
    S_{N_s} = \{ i : \mu_i \in \text{TopK}(\boldsymbol{\mu}, N_s \cdot m) \}
\end{equation}

\subsection{Balanced Clustering Algorithm for Routed Experts}
\label{appendix:clustering}

We employ a constrained balanced K-means algorithm on the activation feature vectors $\mathbf{c}_i$.

\textbf{Centroid Initialization.} We select $N_r$ centroids from remaining neurons with highest activation rates:
\begin{equation}
C = \{ \hat{\mathbf{c}}_1, \dots, \hat{\mathbf{c}}_{N_r} \}
\end{equation}

\textbf{Distance Matrix.} We construct $\mathbf{D} \in \mathbb{R}^{N_r \cdot m \times N_r}$ where:
\begin{equation}
    d_{i,j} = \| \mathbf{c}_i - \hat{\mathbf{c}}_j \|_2 = \sqrt{\sum_{k=1}^q (c_{k,i} - \hat{c}_{k,j})^2}
\end{equation}

\textbf{Equivalence of $L_2$ and Hamming Distance.}
Since $\mathbf{c}_i \in \{0,1\}^q$ are binary, $L_2$ distance relates directly to Hamming distance:
\begin{align}
    \| \mathbf{c}_i - \mathbf{c}_j \|_2^2 
    &= \sum_{k=1}^{q} (c_{k,i} - c_{k,j})^2 = d_H(\mathbf{c}_i, \mathbf{c}_j)
\end{align}
where $d_H$ counts positions where vectors differ. Minimizing $L_2$ is equivalent to minimizing Hamming distance.

\textbf{Balanced Assignment.} The algorithm iteratively solves:
\begin{align}
    \label{eq:con_1_appendix} 
    \min_{T} &\sum_{i,j} T_{i,j} \cdot d_{i,j} \\
    \text{s.t. } &\sum_{i} T_{i,j} = m, \; \sum_{j} T_{i,j} = 1, \; T_{i,j} \geq 0 \nonumber
\end{align}

\textbf{Cluster Update:}
\begin{equation}
    \hat{\mathbf{c}}_{j}^{t+1} = \begin{cases}
        \frac{\sum_{i} T_{i,j}^{t} \cdot \mathbf{c}_i}{\sum_{i} T_{i,j}^{t}}, & \text{if } \sum_{i} T_{i,j}^{t} > 0, \\
        \hat{\mathbf{c}}_{j}^{t}, & \text{otherwise}.
    \end{cases}
\end{equation}

Since this is an unbalanced assignment ($mN_r$ neurons to $N_r$ clusters of size $m$), we extend $\mathbf{D}$ by repeating each column $m$ times, then solve the balanced assignment using the Jonker-Volgenant algorithm~\cite{jonker1988shortest} with complexity $O(n^3)$.

\subsection{Router Construction Derivation}
\label{appendix:router}

This section provides the complete derivation for the analytical router.

\textbf{Problem.} Given input $\mathbf{x}$, the reconstruction error is:
\begin{align}
\label{eq:prob_appendix}
    &\arg\min_{G} | F_{MoE}(\mathbf{x}) - F(\mathbf{x})| \nonumber \\
    &= \arg\min_{G} |\sum_{i \in S_{de}} E_i^r(\mathbf{x})|
\end{align}
where $S_{de}$ is the set of deactivated experts.

\textbf{Reduction.} Using the sparsity hypothesis (\Cref{eq:hypo_appendix}), the reconstruction error can be bounded by the sum of hidden-state magnitudes of deactivated experts:
\begin{align}
\label{eq:reduced_appendix}
    &\arg\min_{G} \left|\sum_{i \in S_{de}} E_i^r(\mathbf{x})\right| \nonumber \\
    &\approx \arg\min_{G} \mathbb{E}_{\mathbf{x}}\left[ \sum_{i \in S_{de}} \|\mathbf{h}^r_i\|_1 \right].
\end{align}

\textbf{Optimal Solution.} The optimal router matches sorting of expert scores $\{s_i\}$ with expected activations $\bar{\mathbf{h}}^r_i = \mathbb{E}[\|\mathbf{h}^r_i\|_1]$:
\begin{align}
\label{eq:permu_appendix}
    s_{\sigma(1)} \leq \cdots \leq s_{\sigma(N_r)}, \quad \bar{\mathbf{h}}^r_{\sigma(1)} \leq \cdots \leq \bar{\mathbf{h}}^r_{\sigma(N_r)}
\end{align}

\textbf{Representative Neuron.} For each expert $j$, the representative neuron is:
\begin{equation}
    R_j = \argmin_{i \in \text{cluster } j} \|\mathbf{c}_i - \hat{\mathbf{c}}_j\|_2.
\end{equation}
The router uses these neurons:
\begin{equation}
G(\mathbf{x}) = \text{Swish}(\mathbf{W}_{\text{gate}}^{R\top} \mathbf{x}) \odot (\mathbf{W}_{\text{up}}^{R\top} \mathbf{x})
\end{equation}
yielding scores that approximate expert contributions.

\end{document}